\documentclass{article}

\usepackage[preprint]{nips}

\usepackage[utf8]{inputenc} 
\usepackage[T1]{fontenc}    
\usepackage{hyperref}       
\usepackage{url}            
\usepackage{booktabs}       
\usepackage{amsfonts}       
\usepackage{nicefrac}       
\usepackage{microtype}      
\usepackage{xcolor}         

\usepackage{graphicx}
\usepackage{amsmath}
\usepackage{amssymb}
\usepackage{booktabs}
\usepackage{float}

\usepackage{bm}
\usepackage{bbm}
\usepackage{mathtools}
\usepackage{amsfonts}       
\usepackage{nicefrac}       
\usepackage{microtype}      
\usepackage{multirow}
\usepackage{adjustbox}
\usepackage{wrapfig}
\usepackage{placeins}
\usepackage[normalem]{ulem}
\useunder{\uline}{\ul}{}

\usepackage{color}
\definecolor{Red7}{rgb}{0.941, 0.243, 0.243}
\definecolor{Green7}{RGB}{55, 178, 77}
\definecolor{Tdgreen}{rgb}{0,0.4,0.7}
\usepackage{pifont}
\usepackage[capitalize]{cleveref}
\crefname{section}{Sec.}{Secs.}
\Crefname{section}{Section}{Sections}
\Crefname{table}{Table}{Tables}
\crefname{table}{Tab.}{Tabs.}

\newfloat{figtab}{htb}{fgtb}
\makeatletter
  \newcommand\figcaption{\def\@captype{figure}\caption}
  \newcommand\tabcaption{\def\@captype{table}\caption}
\makeatother

\title{Segment Any Anomaly without Training via \\  Hybrid Prompt Regularization}

\author{
  Yunkang Cao$^{1}$\footnotemark[1]\quad 
  Xiaohao Xu$^{1}$\footnotemark[1]\quad 
  Chen Sun$^{1}$ \quad 
  \textbf{Yuqi Cheng}$^{1}$ \\
  \textbf{Zongwei Du}$^{1}$ \quad
  \textbf{Liang Gao}$^{1}$  \quad
  \textbf{Weiming Shen$^{1}$\footnotemark[4]} 
  \\
$^1$ State Key Laboratory of Digital Manufacturing 
Equipment and Technology,\\ Huazhong University of Science and Technology, China\\
  \texttt{\{cyk\_hust, sun\_chen, chengyuqi, duzongwei, gaoliang\}@hust.edu.cn}  \\  \texttt{xxh11102019@outlook.com}, \texttt{chengyuqi.c@qq.com},  \texttt{wshen@ieee.org} \\
}

\begin{document}

\maketitle

\renewcommand{\thefootnote}{\fnsymbol{footnote}}
\footnotetext[1]{Equal Contribution.}
\footnotetext[4]{Corresponding Author.}

\renewcommand{\thefootnote}{\arabic{footnote}}

\begin{abstract}
We present a novel framework, \textit{i.e.}, Segment Any Anomaly + (SAA$+$), for zero-shot anomaly segmentation with hybrid prompt regularization to improve the adaptability of modern foundation models. Existing anomaly segmentation models typically rely on domain-specific fine-tuning, limiting their generalization across countless anomaly patterns. In this work, inspired by the great zero-shot generalization ability of foundation models like Segment Anything, we first explore their assembly to leverage diverse multi-modal prior knowledge for anomaly localization. For non-parameter foundation model adaptation to anomaly segmentation, we further introduce hybrid prompts derived from domain expert knowledge and target image context as regularization. Our proposed SAA$+$ model achieves state-of-the-art performance on several anomaly segmentation benchmarks, including VisA, MVTec-AD, MTD, and KSDD2, in the zero-shot setting. We will release the code at \href{https://github.com/caoyunkang/Segment-Any-Anomaly}{https://github.com/caoyunkang/Segment-Any-Anomaly}.

\end{abstract}

\section{Introduction}
\label{sec:intro}
 Anomaly segmentation models~\cite{cao_collaborative_2023, wan_industrial_2022, roth2022towards} have attracted great interest in various domains, \textit{e.g,}, industrial quality control~\cite{bergmann2019mvtec, bergmann2020uninformed} and medical diagnoses~\cite{baur_autoencoders_2021}. The key to reliable anomaly segmentation is to discriminate the distribution of anomaly data from normal data. Specifically, this paper considers zero-shot anomaly segmentation (ZSAS) on images, which is a promising yet unexplored setting where neither normal nor abnormal image is provided for the target category during  training.

Due to the scarcity of abnormal samples for training, many works are working towards unsupervised or self-supervised anomaly segmentation, which targets learning a representation of the normal samples during training. Then, the anomalies can be segmented by calculating the discrepancy between the test sample and the learned normal distribution. In specific, these models, including auto-encoder-based reconstruction~\cite{zhou2020encoding,hou2021divide,zavrtanik2021draem,matsubara2020deep,yan2021learning,jiang2022masked}, one-class classification~\cite{yi2020patch,massoli2021mocca,sohn2020learning}, and memory-based normal distribution~\cite{roth2022towards, wan_industrial_2022, cao2023complementary, wang_multimodal_nodate, jiang_softpatch_2022} methods, typically require training separate models for certain limited categories. However, in real-world scenarios, there are millions of industrial products, and it is not cost-effective to collect a large training set for individual objects, which hinders their deployment in cases when efficient deployments are required, \textit{e.g.}, the initial stage of production.

\begin{figure}[t]
  \centering
  \includegraphics[width=\linewidth]{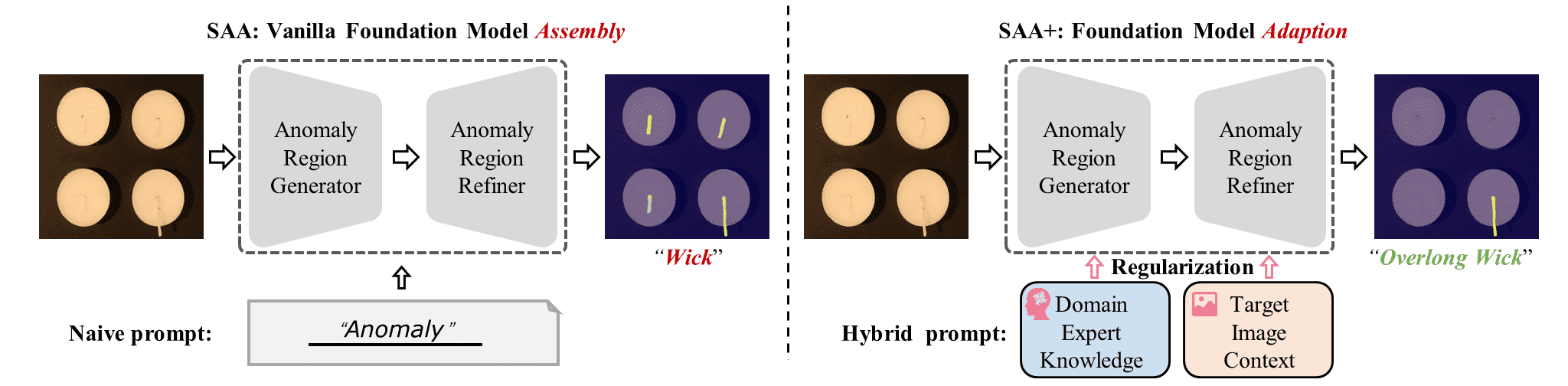}
   \vspace{-5mm}
   \caption{Towards segmenting any anomaly without training, we first construct a vanilla baseline (SAA) by prompting into a cascade of anomaly region generator (\textit{e.g.}, a prompt-guided object detection foundation model~\cite{liu2023grounding}) and  anomaly region refiner (\textit{e.g.}, a segmentation foundation model~\cite{kirillov2023segment}) modules via a naive class-agnostic language prompt (\textit{e.g.}, ``Anomaly''). However, SAA shows the severe false-alarm problem, which falsely detects all the ``\texttt{\textcolor[RGB]{207,63,63}{wick}}'' rather than the ground-truth anomaly region (the ``\texttt{\textcolor[RGB]{112,173,71}{overlong wick}}''). Thus, we further strengthen the regularization with hybrid prompts in the revamped model (SAA$+$), which successfully helps identify the anomaly region.}
  \label{fig:teaser}
  \vspace{-1mm}
\end{figure}

Recently, foundation models, \textit{e.g.}, SAM~\cite{kirillov2023segment} and CLIP~\cite{radford2021learning}, exhibit great zero-shot visual perception abilities by retrieving prior knowledge stored in these models via prompting~\cite{li2022align, bommasani2021opportunities}.  In this work, we would like to explore how to adapt foundation models to realize anomaly segmentation under the zero-shot setting. To this end, as is shown in Fig.~\ref{fig:teaser}, we first construct a vanilla baseline, \textit{i.e.}, Segment Any Anomaly (SAA), by cascading prompt-guided object detection~\cite{liu2023grounding} and segmentation foundation models~\cite{kirillov2023segment}, which serve as Anomaly Region Generator and Anomaly Region Refiner, respectively. Following the practice to unlock foundation model knowledge~\cite{clipseg2022,jeong2023winclip}, naive language prompts, \textit{e.g.}, ``\verb|defect|'' or ``\verb|anomaly|'', are utilized to segment desired anomalies for a target image. In specific, the language prompt is used to prompt the Anomaly Region Generator to generate prompt-conditioned box-level regions for desired anomaly regions. Then these regions are refined in the Anomaly Region Refiner to produce final predictions, \textit{i.e.}, masks, for anomaly segmentation. 

However, as is shown in Figure \ref{fig:teaser}, vanilla foundation model assembly (SAA) tends to cause significant false alarms, \textit{e.g.}, SAA wrongly refers to all wicks as anomalies whereas only the overlong wick is a real anomaly, which we attribute to the \textit{ambiguity} brought by naive language prompts. Firstly, conventional language prompts may become ineffective when facing the domain shift between the pre-training data distribution of foundation models and downstream datasets for anomaly segmentation. Secondly, the degree of ``\verb|anomaly|'' for a target depends on the object context, which is hard for naive coarse-grained language prompts, \textit{e.g.}, ``\verb|an anomaly region|'', to express exactly.


Thus, going beyond naive language prompts, we incorporate domain expert knowledge and target image context in our revamped framework, \textit{i.e.}, Segment Any Anomaly + (SAA$+$), respectively.
On the one hand, expert knowledge provides detailed descriptions of anomalies that are relevant to the target in open-world scenarios. We utilize more specific descriptions as in-context prompts, effectively aligning the image content in both pre-trained and target datasets. On the other hand, we utilize the target image context to reliably identify and adaptively calibrate anomaly segmentation predictions~\cite{object_calibration, xu2022reliable}. By leveraging the rich contextual information present in the target image, we can accurately associate the object context with the final anomaly predictions. 

Technically, apart from naive class-agnostic prompts, we leverage domain expert knowledge to construct target-oriented anomaly language prompts, \textit{i.e.}, class-specific language expressions. Besides, as language can not accurately retrieve regions with certain object characteristics, such as number, size, and location, precisely~\cite{paiss_count_2023,li2022r}, we introduce object property prompts in the form of thresholding filters. These prompts assist in identifying and removing region candidates that do not satisfy desired properties. Furthermore, to fully exploit the target image context, we suggest utilizing image saliency and region confidence ranking as prompts, which model the anomaly degree of a region by considering the similarities, \textit{e.g.}, euclidean distance, between it and other regions within the image. Finally, we conduct thorough experiments to confirm the efficacy of our hybrid prompts in adapting foundation models to zero-shot anomaly segmentation. Specifically, our final model (SAA$+$) attains new state-of-the-art performance on various anomaly segmentation datasets under the zero-shot setting.
To summarize, our main contributions are:
\begin{itemize}
    \item We propose the SAA framework for anomaly segmentation, allowing the collaborative assembly of diverse foundation models without the need for training.
    \item We introduce hybrid prompts as a regularization technique, leveraging domain expert knowledge and target image context to adapt foundation models for anomaly segmentation. This leads to the development of SAA$+$, an enhanced version of our framework.
    \item Our method achieves state-of-the-art performance in zero-shot anomaly segmentation on several benchmark datasets, including VisA, MVTec-AD, KSDD2, and MTD. Notably, SAA/SAA$+$ demonstrates remarkable capability in detecting texture-related anomalies without requiring any annotation.
\end{itemize}


\section{Related work}
\label{sec:related_work}

\noindent\textbf{Anomaly Segmentation. } Due to the limited availability and high cost of abnormal images in industrial settings, much of the current research on anomaly segmentation focuses on unsupervised methods that rely solely on normal images. Reconstruction-based approaches, such as those proposed in \cite{zhou2020encoding,hou2021divide,zavrtanik2021draem,matsubara2020deep,yan2021learning,jiang2022masked}, score anomalies with 
train an encoder-decoder model to reconstruct images for segmentation purposes. By comparing the input image with the reconstructed version, these methods can predict the location of anomalies. Feature embedding-based methods, on the other hand, typically employ teacher-student architecture \cite{salehi2021multiresolution,wang2021student,deng2022anomaly,cao2022informative, cao2022semikd, wan_unsupervised_2022, wan_position_2022, cao_collaborative_2023}, one-class classification technology \cite{yi2020patch,massoli2021mocca,sohn2020learning}, or memory-based normal distribution~\cite{roth2022towards, wan_industrial_2022, cao2023complementary} to segment anomalies by identifying differences in feature distribution between normal and abnormal images. 

Recently, researchers have begun to explore the potential of ZSAS~\cite{nagy2022zero,liu2021zero,rivera2020anomaly,aota2023zero}, which eliminates the need for either normal or abnormal images during the training process. Among them, WinClip~\cite{jeong2023winclip} pioneers the potential of foundation models, \textit{e.g.}, visual-language models, for the ZSAL task. Unlike WinClip~\cite{jeong2023winclip} that segments anomalies through text-visual similarity, we propose to generate proposals and score their anomaly degree, achieving much better segmentation performance.

\noindent\textbf{Foundation Model.} 
Foundation models show an impressive ability to solve diverse vision tasks in a zero-shot manner. Specifically, these models can learn a strong representation by training on large-scale datasets ~\cite{Laion400}. While early work~\cite{radford2021learning,li2021align} focus on developing robust image-wise recognition capacity, recent work~\cite{lu2022unified,wang2022unifying, rao2022denseclip,zhong2022regionclip,zhou2021denseclip,liu2023grounding} introduce foundation models or their applications for dense visual tasks. For instance, Grounding DINO \cite{liu2023grounding} achieves encouraging open-set object detection ability using arbitrary texts as queries. Recently, SAM \cite{kirillov2023segment} demonstrates a powerful ability to extract high-quality object segmentation masks in the open world. Impressed by the success of these foundation models, we would like to explore how to adapt these off-the-shelf models to detect anomalies without any training on the downstream datasets for anomaly segmentation.

\noindent\textbf{Prompt Engineering. } 
Prompt engineering is a widely employed technique that involves adapting foundation models for downstream tasks. Generally, this approach involves appending a set of learnable tokens to the input.  Prior studies have investigated prompting with text inputs~\cite{zhou_conditional}, vision inputs~\cite{ju_prompting_2022, jia_visual_2022, bahng_exploring_2022}, and both text and visual inputs~\cite{zang_unified_2022, shen_multitask_2022, zhou_learning_2022}. Despite their effectiveness in adapting foundation models to various downstream tasks, prompting methods cannot be employed in ZSAS because they require training data, which is not available in ZSAS. In contrast, some methods employ heuristic prompts~\cite{shtedritski_what_2023} that do not require any training, making them more feasible for tasks without any data. In this paper, we propose using hybrid prompts derived from domain expert knowledge and target image context for ZSAS. 

\begin{figure*}[!t]
  \centering
  \includegraphics[width=\linewidth]{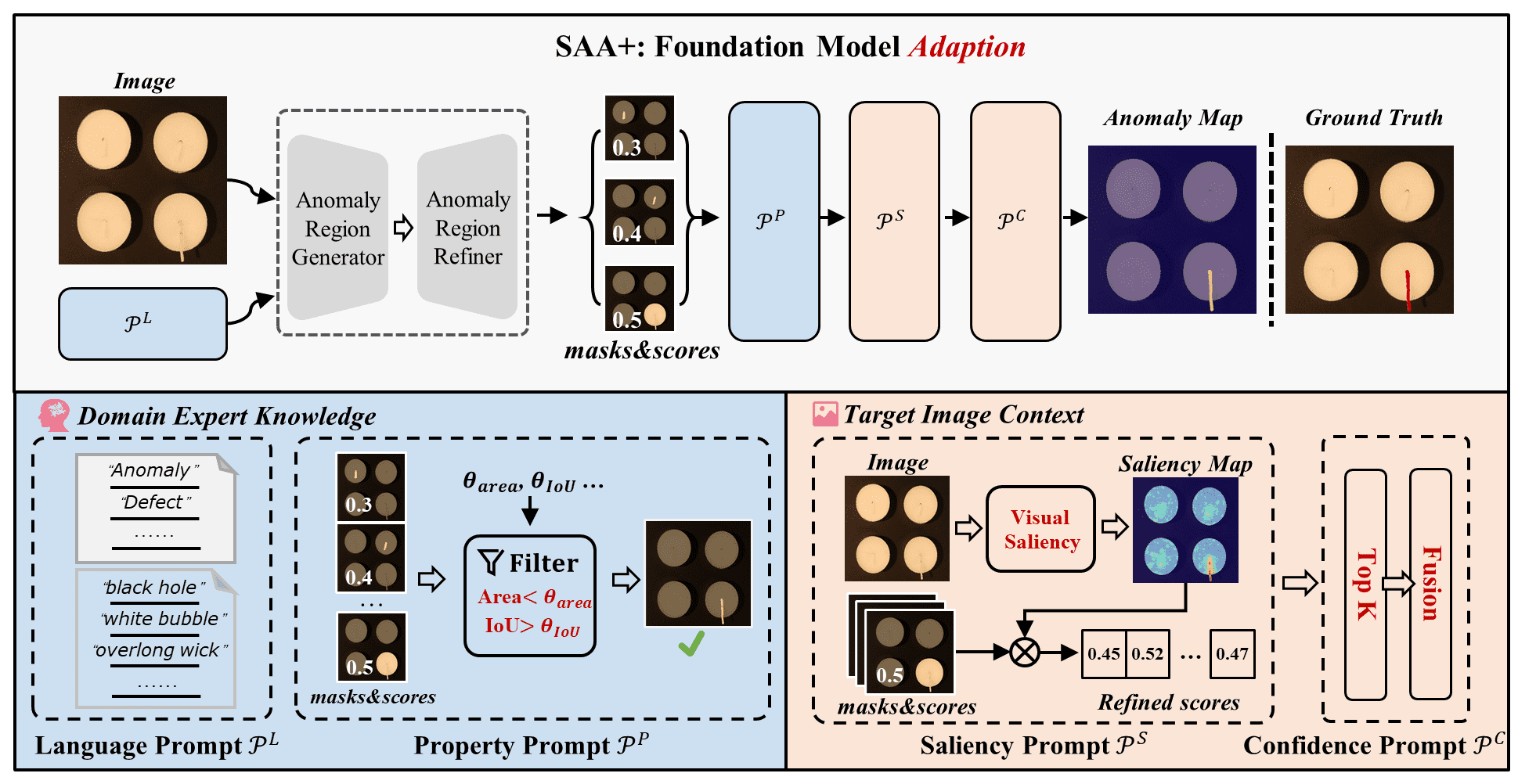}
  \vspace{-0.15in}
  \caption{\textbf{Overview of the proposed Segment Any Anomaly + (SAA+) framework.} We adapt foundation models to zero-shot anomaly segmentation via hybrid prompt regularization. In specific, apart from naive class-agnostic language prompts, the regularization comes from both domain expert knowledge, including more detailed class-specific language and object property prompts, and target image context, including visual saliency and confidence ranking-related prompts.
  }  
  \label{fig:framework}
  
\end{figure*}

\section{SAA: Vanilla Foundation Model Assembly for ZSAS}

\subsection{Problem Definition: Zero-shot Anomaly Segmentation (ZSAS)}

The goal of ZSAS is to perform anomaly segmentation on new objects without requiring any corresponding object training data. ZSAS seeks to create an anomaly map $\mathbf{A} \in [0,1]^{h \times w \times 1}$ based on an empty training set $\emptyset$, in order to identify the anomaly degree for individual pixels in an image $\mathbf{I} \in \mathbb{R}^{h \times w \times 3}$ that includes novel objects. The ZSAS task has the potential to significantly reduce the need for training data and lower the costs associated with real-world inspection deployments.

\subsection{Baseline Model Assembly: Segment Any Anomaly (SAA) }

For ZSAS, we start by constructing a vanilla foundation model assembly, \textit{i.e.}, Segment Any Anomaly (SAA), as shown in Fig. \ref{fig:teaser}. In specific, given a certain query image for anomaly segmentation, we first use languages as the initial prompt to roughly retrieve coarse anomaly region proposals via an Anomaly Region Generator implemented with a language-driven visual grounding foundation model, \textit{i.e.}, GroundingDINO~\cite{liu2023grounding}. Afterward, anomaly region proposals are refined into pixel-wise high-quality segmentation masks with the Anomaly Region Refiner in which a prompt-driven segmentation foundation model, \textit{i.e.}, SAM~\cite{kirillov2023segment}, is used.

\subsubsection{Anomaly Region Generator}

With recent booming development on language-vision models, some foundation models~\cite{clipseg2022,liu2023grounding,zhong2022regionclip} gradually acquire the ability to retrieve objects in images through language prompts. Given language prompts $\mathcal{T}$ that describe desired regions to be detected, \textit{e.g.}, ``\verb|anomaly|'', foundation models can generate desired regions for  a query image $\mathbf{I}$. There we base the architecture of the region detector on a text-guided open-set object detection architecture for visual grounding. Specifically, we take a GroundingDINO~\cite{liu2023grounding} architecture that has been pre-trained on large-scale language-vision datasets~\cite{Laion400}. Such a network first extracts the features of the language prompt and the query image via text encoder and visual encoder, respectively. Then the rough object regions are generated in the form of bounding boxes with a cross-modality decoder. Given the bounding-box-level region set $\mathcal{R}^B$, and their corresponding confidence score set $\mathcal{S}$, the module of anomaly region generator ($\mathrm{Generator}$) can be formulated as,
\begin{equation}
\label{eq:det}
    \mathcal{R}^B, \mathcal{S} := \mathrm{Generator}(\mathbf{I},\mathcal{T})
\end{equation}

\subsubsection{Anomaly Region Refiner}

To generate pixel-wise anomaly segmentation results, we propose Anomaly Region Refiner to refine the bounding-box-level anomaly region candidates into an anomaly segmentation mask set. To this end, we use a sophisticated foundation model for open-world visual segmentation, \textit{i.e.}, SAM~\cite{kirillov2023segment}. This model mainly includes a ViT-based~\cite{dosovitskiy2021an} backbone and a prompt-conditioned mask decoder. In specific, the model is trained on a large-scale image segmentation dataset~\cite{kirillov2023segment} with one billion fine-grained masks, which enables high-quality mask generation abilities under an open-set segmentation setting. The prompt-conditioned mask decoder accepts various types of prompts as input. We regard the bounding box candidates $\mathcal{R}^B$ as prompts and obtain pixel-wise segmentation masks $\mathcal{R}$. The module of the Anomaly Region Refiner ($\mathrm{Refiner}$) can be formulated as follows,
\begin{equation}
    \mathcal{R} := \mathrm{Refiner}(\mathbf{I}, \mathcal{R}^B)
\end{equation}

Till then, we obtain the set of regions in the form of high-quality segmentation masks $\mathcal{R}$ with corresponding confidence scores $\mathcal{S}$. To sum up we summarize framework ($\mathrm{SAA}$) as follows,
\begin{equation}
    \mathcal{R}, \mathcal{S} := \mathrm{SAA}(\mathbf{I}, \mathcal{T}_n)
\end{equation}
where $\mathcal{T}_n$ is a naive class-agnostic language prompt, \textit{e.g.}, ``\verb|anomaly|'', utilized in SAA. 

\subsection{Analysis on the ZSAS Performance of Vanilla Foundation Model Assembly}

We present some preliminary experiments to evaluate the efficacy of vanilla foundation model assembly for ZSAS. Despite the simplicity and intuitiveness of the solution, we observe a \textit{language ambiguity} issue. Specifically, certain language prompts, such as ``anomaly'', may fail to detect the desired anomaly regions. For instance, as depicted in Fig. \ref{fig:teaser}, all ``\verb|wick|'' is erroneously identified as an anomaly by the SAA with the ``\verb|anomaly|'' prompt. 

We attribute this language ambiguity to the domain gap between the pretraining language-vision datasets and the targeted ZSAS datasets, which means that some language prompts may have different meanings and be associated with different image contents in distinct datasets. In addition, there is hardly any adjective expression like ``\verb|anomaly|'' in those large-scale datasets, thus making this kind of prompt design poor at understanding what is an anomaly region. Additionally, the exact ``\verb|anomaly|'' is object-specific and would vary across objects. For example, it denotes the scratches on leather or the crack on hazelnut. The language ambiguity issue leads to severe false alarms in ZSAS datasets. We propose introducing hybrid prompts generated by domain expert knowledge and the target image context to reduce language ambiguity, thereby achieving better ZSAS performance. 

\section{SAA+: Foundation Model Adaption via Hybrid Prompt Regularization}
\label{sec:method}  

To address language ambiguity in SAA and improve its ability on ZSAS, we propose an upgraded version called SAA$+$ that incorporates hybrid prompts, as Fig. \ref{fig:framework}. In addition to leveraging the knowledge gained from pre-trained foundation models, SAA$+$ utilizes both domain expert knowledge and target image context to generate more accurate anomaly region masks. We provide further details on these hybrid prompts below.

\subsection{Prompt Generated from Domain Expert Knowledge}

Following the trend of prompt learning~\cite{zhou_conditional,zhou_learning_2022}, we initialize the prompt, which unlocks the knowledge of foundation models, in the form of language. However, the language ambiguity issue caused by the domain gap is particularly severe when using only the naive language prompt ``\verb|anomaly|''. To address this problem, we leverage domain expert knowledge that contains useful prior information about the target anomaly regions. Specifically, although experts may not provide a comprehensive list of potential open-world anomalies for a new product, they can identify some candidates based on their past experiences with similar products. Domain expert knowledge enables us to refine the naive ``\verb|anomaly|'' prompt into more specific prompts that describe the anomaly state in greater detail. In addition to language prompts, we introduce property prompts to complement the lack of awareness on specific properties like ``\verb|count|'' and ``\verb|area|'' ~\cite{paiss_count_2023} in existing foundation models~\cite{paiss_count_2023}.

\subsubsection{Anomaly Language Expression as Prompt}

To describe potential open-world anomalies, we propose designing more precise language prompts. These prompts are categorized into two types: class-agnostic and class-specific prompts. 

\noindent \textbf{Class-agnostic prompts ($\mathcal{T}_{\rm a}$}) are general prompts that describe anomalies that are not specific to any particular category, \textit{e.g.}, ``\verb|anomaly|'' and ``\verb|defect|''. Despite the domain gap between the pre-trained datasets and the targeted ZSAS datasets, our empirical analysis (\ref{sec:ablation}) shows that these generic prompts provide  encouraging initial performance.

\noindent \textbf{Class-specific prompts ($\mathcal{T}_{\rm s}$}) are designed based on expert knowledge of abnormal patterns with similar products to supplement more specific anomaly details.
We use prompts already employed in the pre-trained visual-linguistic dataset, \textit{e.g.}, ``\verb|black hole|'' and ``\verb|white bubble|'', to query the desired regions. This approach reformulates the task of finding an anomaly region into locating objects with a specific anomaly state expression, which is more straightforward to utilize foundation models than identifying ``\verb|anomaly|'' within an object context.

By prompting SAA with anomaly language prompts $\mathcal{P}^L=\{ \mathcal{T}_{\rm a}, \mathcal{T}_ {\rm s}   \}$ derived from domain expert knowledge, we generate finer anomaly region candidates $\mathcal{R}$ and corresponding confidence scores $\mathcal{S}$. 


\subsubsection{Anomaly Object Property as Prompt}

Current foundation models~\cite{liu2023grounding, li2021grounded} have limitations when it comes to querying objects with specific property descriptions, such as size or location, which are important for describing anomalies, such as ``\verb|The small black hole on the left of the cable.|'' To incorporate this critical expert knowledge, we propose using anomaly property prompts formulated as rules rather than language. Specifically, we consider the location and area of anomalies.

\noindent\textbf{Anomaly Location.} Accurate localization of anomalies plays a critical role in distinguishing true anomalies from false positives. Typically, anomalies are expected to be located within the objects of interest during inference. However, due to the influence of background context, anomalies may occasionally appear outside the inspected objects. To tackle this challenge, we leverage the open-world detection capability of foundation models to determine the location of the inspected object. Subsequently, we calculate the intersection over union (IoU) between the potential anomaly regions and the inspected object. By applying an expert-derived IoU threshold, denoted as $\theta_{IoU}$, we filter out anomaly candidates with IoU values below this threshold. This process ensures that the retained anomaly candidates are more likely to represent true anomalies located within the inspected object.

\noindent\textbf{Anomaly Area.} The size of an anomaly, as reflected by its area, is also a property that can provide useful information. In general, anomalies should be smaller than the size of the inspected object. Experts can provide a suitable threshold value $\theta_{area}$ for the specific type of anomaly being considered. Candidates with areas unmatched with $\theta_{area} \cdot \mathrm{Object Area}$ can then be filtered out.

By combining the two property prompts $\mathcal{P}^P=\{ \theta_{area}, \theta_{IoU} \}$, we can filter the set of candidate regions $\mathcal{R}$ to obtain a subset of selected candidates $\mathcal{R}^P$ with corresponding confidence scores $\mathcal{S}^P$ using the filter function ($\mathrm{Filter}$),
\begin{equation}
    \mathcal{R}^P, \mathcal{S}^P := \mathrm{Filter}(\mathcal{R}, \mathcal{P}^P)
\end{equation}

\subsection{Prompts Derived from Target Image Context}

Besides incorporating domain expert knowledge, we can leverage the information provided by the input image itself to improve the accuracy of anomaly region detection. In this regard, we propose two prompts induced by the image context.

\subsubsection{Anomaly Saliency as Prompt}
Predictions generated by foundation models like~\cite{liu2023grounding} using the prompt ``\verb|defect|'' can be unreliable due to the domain gap between pre-trained language-vision datasets~\cite{Laion400} and targeted anomaly segmentation datasets~\cite{bergmann2019mvtec,zou2022spot}. To calibrate the confidence scores of individual predictions, we propose Anomaly Saliency Prompt mimicking human intuition. In specific, humans can recognize anomaly regions by their discrepancy with their surrounding regions \cite{aota2023zero}, \textit{i.e.}, visual saliency contains valuable information indicating the anomaly degree. Hence, we calculate a saliency map ($\mathbf{s}$) for the input image by computing the average distances between the corresponding pixel feature ($\mathbf{f}$) and its $N$ nearest neighbors, 
\begin{equation}
\label{eq:saliency-map}
    \mathbf{s}_{ij} := \frac{1}{N}\sum\limits_{\mathbf{f}\in N_p(\mathbf{f}_{ij})}(1-  \langle \mathbf{f}_{ij},\mathbf{f}  \rangle)
\end{equation}
where $(i,j)$ denotes to the pixel location, $N_p(\mathbf{f}_{ij})$ denotes to the $N$ nearest neighbors of the corresponding pixel, and $\langle \cdot, \cdot \rangle$ refers to the cosine similarity. We use pre-trained CNNs from large-scale image datasets~\cite{hinton2012imagenet} to extract image features, ensuring the descriptiveness of features. The saliency map indicates how different a region is from other regions. The saliency prompts $\mathcal{P}^{S}$ are defined as the exponential average saliency value within the corresponding region masks, 
\begin{equation}
\label{eq:score_saliency}
    \mathcal{P}^{S} := \left\{ \exp(\frac{\sum_{i j}\mathbf{r}_{i j}\mathbf{s}_{i j}}{\sum_{i j}\mathbf{r}_{i j}}) \quad | \quad \mathbf{r} \in \mathcal{R}^P \right\}
\end{equation}

The saliency prompts provide reliable indications of the confidence of anomaly regions. These prompts are employed to recalibrate the confidence scores generated by the foundation models, yielding new rescaled scores $\mathcal{S}^{S}$ based on the anomaly saliency prompts $\mathcal{P}^{S}$. These rescaled scores provide a combined measure that takes into account both the confidence derived from the foundation models and the saliency of the region candidate. The process is formulated as follows,
\begin{equation}
\label{eq:rescore}
    \mathcal{S}^{S} := \left\{ p \cdot s  \quad | \quad   p \in \mathcal{P}^{S},   s \in \mathcal{S}^P \right\}
\end{equation}

\subsubsection{Anomaly Confidence as Prompt}

Typically, the number of anomaly regions in an inspected object is limited. Therefore, we propose anomaly confidence prompts $\mathcal{P}^C$ to identify the $K$ candidates with the highest confidence scores based on the image content and use their average values for final anomaly region detection. This is achieved by selecting the top $K$ candidate regions based on their corresponding confidence scores, as shown in the following,
\begin{equation}
\label{eq:confidence-prompts}
    \mathcal{R}^C, \mathcal{S}^C := \mathrm{Top}_K(\mathcal{R}^P,\mathcal{S}^S)
\end{equation}

Denote a single region and its corresponding score as $\mathbf{r}^C$ and $s^C$, we then use these $K$ candidate regions to estimate the final anomaly map, 
\begin{equation}
\label{eq:fusion}
    \mathbf{A}_{ij} := \frac{ \sum_{\mathbf{r}^C \in \mathcal{R}^C}
    \mathbf{r}^C_{i j} \cdot s^C}{
    \sum_{\mathbf{r}^C \in \mathcal{R}^C}
    \mathbf{r}^C_{i j}}   
\end{equation}

With the proposed hybrid prompts ($ \mathcal{P}^L, \mathcal{P}^P,\mathcal{P}^S$, and $\mathcal{P}^C$), SAA is regularized in our final framework, \textit{i.e.}, Segment Any Anomaly $+$ (SAA$+$), which makes more reliable anomaly predictions.


\section{Experiments}
\label{sec:experiment}
In this section, we first assess the performance of SAA/SAA$+$ on several anomaly segmentation benchmarks. Then, we extensively study the effectiveness of individual hybrid prompts.

\subsection{Experimental Setup}

\noindent\textbf{Datasets.} We leverage four datasets with pixel-level annotations.: VisA~\cite{zou2022spot}, MVTec-AD~\cite{bergmann2019mvtec}, KSDD2~\cite{bozic_mixed_2021}, and MTD~\cite{mtd2018}. VisA and MVTec-AD comprise a variety of object subsets, \textit{e.g.}, circuit boards, while KSDD2 and MTD are comprised of texture anomalies. 
In summary, we categorize the subsets of all of these datasets into \textit{texture} which typically exhibit similar patterns within a single image (\textit{e.g.}, carpets), and \textit{object}  which includes more diverse distribution (\textit{e.g.}, candles).

\noindent\textbf{Evaluation Metrics.} ZSAS performance is evaluated based on two metrics: 
\noindent (I) \textbf{max-F1-pixel} ($\mathcal{F}_{p}$) ~\cite{jeong2023winclip}, which measures the F1-score for pixel-wise segmentation at the optimal threshold; 
\noindent (II) \textbf{max-F1-region} ($\mathcal{F}_{r}$), which is proposed in this paper to mitigate the bias towards large defects observed with max-F1-pixel~\cite{bergmann2019mvtec}. Specifically, we compute the F1-score for region-wise segmentation at the optimal threshold, considering a prediction positive if the overlapping value exceeds $0.6$.

\vspace{0.05in}

\noindent\textbf{Implementation Details.} We adopt the official implementations of GroundingDINO\footnote{\url{https://github.com/IDEA-Research/GroundingDINO}} and Segment Anything Model\footnote{\url{https://github.com/facebookresearch/segment-anything}} to construct the vanilla baseline (SAA). Details about the prompts derived from domain expert knowledge are explained in the supplementary material. For the saliency prompts induced from image content, we utilize the WideResNet50~\cite{zagoruyko2016wideresnet} network, pre-trained on ImageNet~\cite{hinton2012imagenet}, and set $N=400$ in line with prior studies~\cite{aota2023zero}. For anomaly confidence prompts, we set the hyperparameter $K$ as $5$ by default. Input images are fixed at a resolution of $400 \times 400$ for evaluation.

\subsection{Main Results}
\label{sec:exp_main}

\noindent\textbf{Methods for Comparison.}  We compare our final model, \textit{i.e.}, Segment Any Anomaly + (SAA$+$) with several concurrent state-of-the-art methods, including WinClip~\cite{jeong2023winclip}, UTAD~\cite{aota2023zero}, ClipSeg~\cite{clipseg2022}, and our vanilla baseline (SAA). For WinClip, we report its official results on VisA and MVTec-AD. For the other three methods, we use official implementations and adapt them to the ZSAS task. Notably, as all methods require no training process, their performance is stable with a variance of $\pm0.00$. 

\begin{table}[t]
\centering
\caption{Qualitative comparisons between SAA$+$ and other concurrent methods on zero-shot anomaly segmentation. Best scores are highlighted in \textbf{bold}. The second best scores are also {\ul{underlined}}.}
\label{tab:exp_main}
\setlength{\tabcolsep}{10.pt}
\resizebox{1\textwidth}{!}{
\begin{tabular}{c|l|cccc|cc|c} 
\toprule
\multirow{2}{*}{Metric} &
  \multirow{2}{*}{Method} &
  \multicolumn{4}{c|}{Per Dataset} &
  \multicolumn{2}{c|}{Per Defect   Type} &
  \multirow{2}{*}{Total} \\ \cmidrule(lr){3-6} \cmidrule(lr){7-8}
 &
   &
  \multicolumn{1}{c|}{VisA} &
  \multicolumn{1}{c|}{MVTec-AD} &
  \multicolumn{1}{c|}{KSDD2} &
  MTD &
  \multicolumn{1}{c|}{Texture} &
  Object &
   \\ \midrule
\multirow{3}[10]{*}{$\mathcal{F}_{p}$} &
  WinClip~\cite{jeong2023winclip} &
  \multicolumn{1}{c|}{{\ul {14.82}}} &
  \multicolumn{1}{c|}{{\ul {31.65}}} &
  \multicolumn{1}{c|}{-} &
  - &
  \multicolumn{1}{c|}{-} &
  {\ul {20.93}} &
  - \\ 
 &
  ClipSeg~\cite{clipseg2022} &
  \multicolumn{1}{c|}{14.32} &
  \multicolumn{1}{c|}{25.42} &
  \multicolumn{1}{c|}{{\ul {34.27}}} &
  9.39 &
  \multicolumn{1}{c|}{27.75} &
  18.30 &
  {\ul {20.58}} \\ 
 &
  UTAD~\cite{aota2023zero} &
  \multicolumn{1}{c|}{6.95} &
  \multicolumn{1}{c|}{23.48} &
  \multicolumn{1}{c|}{22.53} &
  11.37 &
  \multicolumn{1}{c|}{{\ul {29.13}}} &
  12.07 &
  16.19 \\ \cmidrule(l){2-9} 
 &
  SAA &
  \multicolumn{1}{c|}{12.76} &
  \multicolumn{1}{c|}{23.44} &
  \multicolumn{1}{c|}{8.79} &
  {\ul {14.78}} &
  \multicolumn{1}{c|}{20.94} &
  17.35 &
  18.22 \\ 
 &
  SAA$+$ &
  \multicolumn{1}{c|}{\textbf{27.07}} &
  \multicolumn{1}{c|}{\textbf{39.40}} &
  \multicolumn{1}{c|}{\textbf{59.19}} &
  \textbf{35.40} &
  \multicolumn{1}{c|}{\textbf{53.79}} &
  \textbf{28.82} &
  \textbf{34.85} \\ \midrule\midrule
\multirow{3}[10]{*}{$\mathcal{F}_{r}$} &
  ClipSeg~\cite{clipseg2022} &
  \multicolumn{1}{c|}{{\ul {5.65}}} &
  \multicolumn{1}{c|}{19.68} &
  \multicolumn{1}{c|}{9.05} &
  6.55 &
  \multicolumn{1}{c|}{21.37} &
  10.41 &
  13.06 \\ 
 &
  UTAD~\cite{aota2023zero} &
  \multicolumn{1}{c|}{5.32} &
  \multicolumn{1}{c|}{17.53} &
  \multicolumn{1}{c|}{3.56} &
  2.95 &
  \multicolumn{1}{c|}{16.38} &
  9.94 &
  11.49 \\ \cmidrule(l){2-9} 
 &
  SAA &
  \multicolumn{1}{c|}{4.83} &
  \multicolumn{1}{c|}{{\ul {32.49}}} &
  \multicolumn{1}{c|}{{\ul {16.40}}} &
  {\ul {10.63}} &
  \multicolumn{1}{c|}{{\ul {40.31}}} &
  {\ul {13.19}} &
  {\ul {19.74}} \\
 &
  SAA$+$ &
  \multicolumn{1}{c|}{\textbf{14.46}} &
  \multicolumn{1}{c|}{\textbf{49.67}} &
  \multicolumn{1}{c|}{\textbf{39.34}} &
  \textbf{30.27} &
  \multicolumn{1}{c|}{\textbf{60.40}} &
  \textbf{25.70} &
  \textbf{34.07} \\ \bottomrule
\end{tabular}}
\end{table}

\noindent\textbf{Quantitative Results}: As is shown in Table \ref{tab:exp_main}, SAA$+$ method outperforms other methods in both $\mathcal{F}_{p}$ and $\mathcal{F}_{r}$ by a significant margin. Although WinClip~\cite{jeong2023winclip}, ClipSeg~\cite{clipseg2022}, and SAA also use foundation models, SAA$+$ better unleash the capacity of foundation models and adapts them to tackle ZSAS. The remarkable performance of SAA$+$ meets the expectation to segment any anomaly without training.

\begin{figure*}[t]
\vspace{-2mm}
    \centering
        \includegraphics[width=\linewidth]{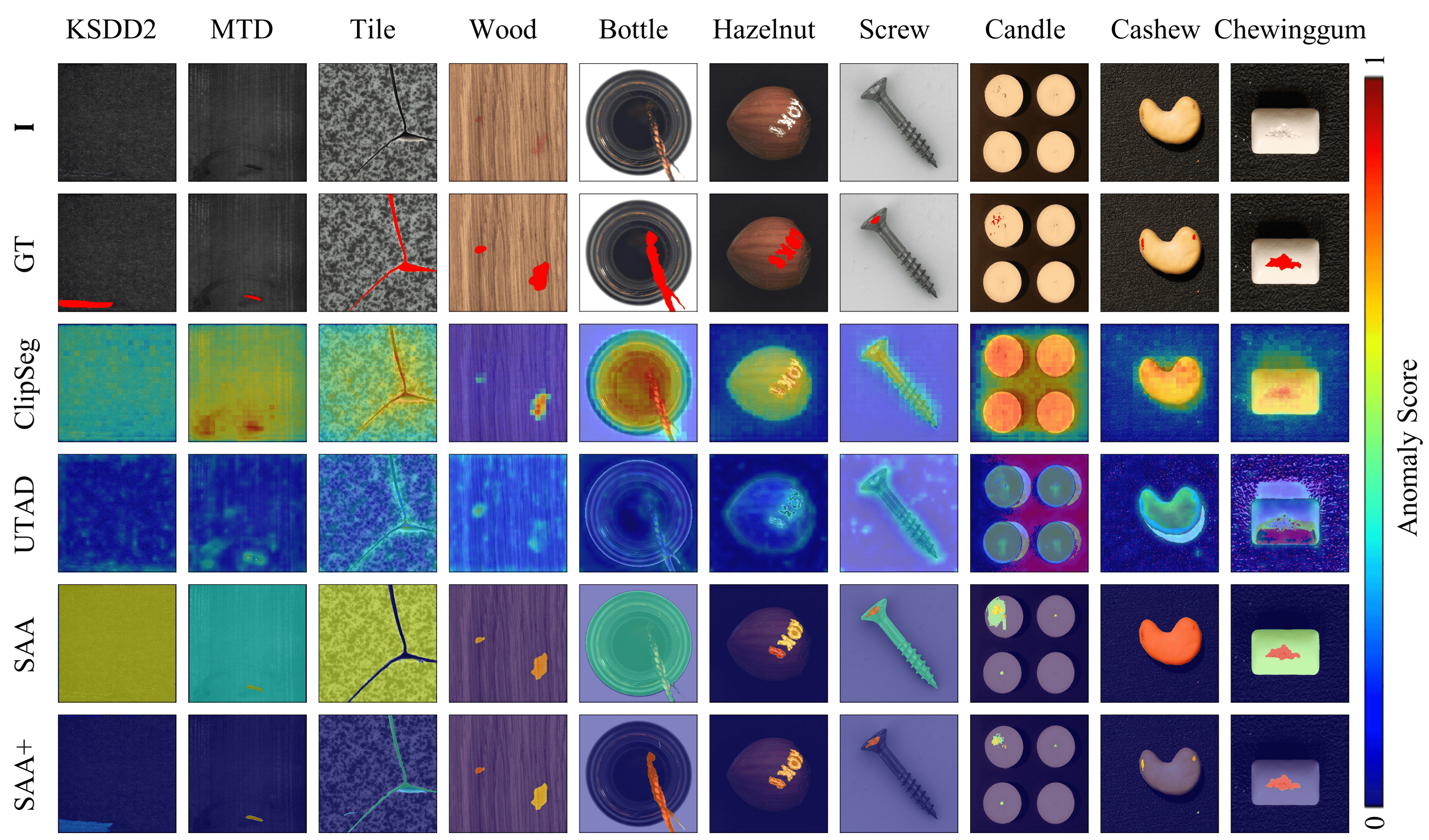}
    \vspace{-4mm}\caption{Qualitative comparisons on zero-shot anomaly segmentation  for ClipSeg~\cite{clipseg2022}, UTAD~\cite{aota2023zero}, SAA, and SAA$+$ on four datasets, \textit{i.e.}, VisA~\cite{zou2022spot}, MVTec-AD~\cite{bergmann2019mvtec}, KSDD2~\cite{bozic_mixed_2021}, and MTD~\cite{mtd2018}}
    \label{fig:qualitative}
\end{figure*}

\noindent\textbf{Qualitative Results}: Fig. \ref{fig:qualitative} presents qualitative comparisons between  SAA$+$ and previous competitive methods, where SAA$+$ achieves better performance. Moreover, the visualization shows SAA$+$ is capable of detecting texture anomalies, \textit{e.g.} small scratches on the leather. 

\begin{figtab}
\begin{minipage}{0.48\linewidth}
    \centering
    \tabcaption{Ablation study on the proposed hybrid prompts, including language prompt ($\mathcal{P}^{L}$), object property prompt ($\mathcal{P}^{P}$), saliency prompt ($\mathcal{P}^{S}$), and confidence prompt ($\mathcal{P}^{C}$). The best scores are highlighted in \textbf{bold}.}
    \vspace{4.4mm}
    \resizebox{\textwidth}{!}{
    \setlength{\tabcolsep}{1pt}
\begin{tabular}{ccllll}
\toprule
\multicolumn{1}{c}{Metric} & \multicolumn{2}{c}{Model Variants}                & Texture        & Object         & Total          \\ \midrule
\multirow{7}[10]{*}{$\mathcal{F}_p$}  & \multirow{3}{*}{w/o $\mathcal{P}^L$} & w/o $\mathcal{T}_{\rm a}$ \& $\mathcal{T}_{\rm s}$ & 50.30          & 24.79          & 30.95          \\
                     &                         & w/o $\mathcal{T}_{\rm a}$       & 51.15          & 25.88          & 31.80          \\
                     &                         & w/o $\mathcal{T}_{\rm s}$       & 53.51          & 26.55          & 33.06          \\ \cmidrule(){2-6}
                     & \multicolumn{2}{c}{w/o $\mathcal{P}^P$}             & 21.83         & 21.40          & 21.50          \\ \cmidrule(){2-6}
                     & \multicolumn{2}{c}{w/o $\mathcal{P}^S$}             & 50.58          & 24.72          & 30.96          \\ \cmidrule(){2-6}
                     & \multicolumn{2}{c}{w/o $\mathcal{P}^C$}             & 50.41          & 27.99          & 34.13          \\ \cmidrule(){2-6}
                     & \multicolumn{2}{c}{full model (\textbf{SAA+})}               & \textbf{53.79} & \textbf{28.82} & \textbf{34.85} \\ \midrule \midrule
\multirow{7}[10]{*}{$\mathcal{F}_r$}  & \multirow{3}{*}{w/o $\mathcal{P}^L$} & w/o $\mathcal{T}_{\rm a}$ \& $\mathcal{T}_{\rm s}$ & 50.58          & 22.36          & 29.17          \\
                     &                         & w/o $\mathcal{T}_{\rm a}$       & 55.26          & 20.28          & 28.72          \\
                     &                         & w/o $\mathcal{T}_{\rm s}$       & 54.21          & 23.13          & 30.64          \\ \cmidrule(){2-6}
                     & \multicolumn{2}{c}{w/o $\mathcal{P}^P$}             & 33.94          & 20.99          & 24.11          \\ \cmidrule(){2-6}
                     & \multicolumn{2}{c}{w/o $\mathcal{P}^S$}             & 57.66          & 24.36          & 32.39          \\ \cmidrule(){2-6}
                     & \multicolumn{2}{c}{w/o $\mathcal{P}^C$}             & 53.65          & 25.18          & 32.05          \\ \cmidrule(){2-6}
                     & \multicolumn{2}{c}{full model (\textbf{SAA+})}               & \textbf{60.40} & \textbf{25.70} & \textbf{34.07} \\ \bottomrule
\end{tabular}}\label{tab:ablation}
  \end{minipage}
  \hfill
  \begin{minipage}[t]{0.48\linewidth}
      \vspace{-50mm}
    \centering
    \includegraphics[width = \linewidth]{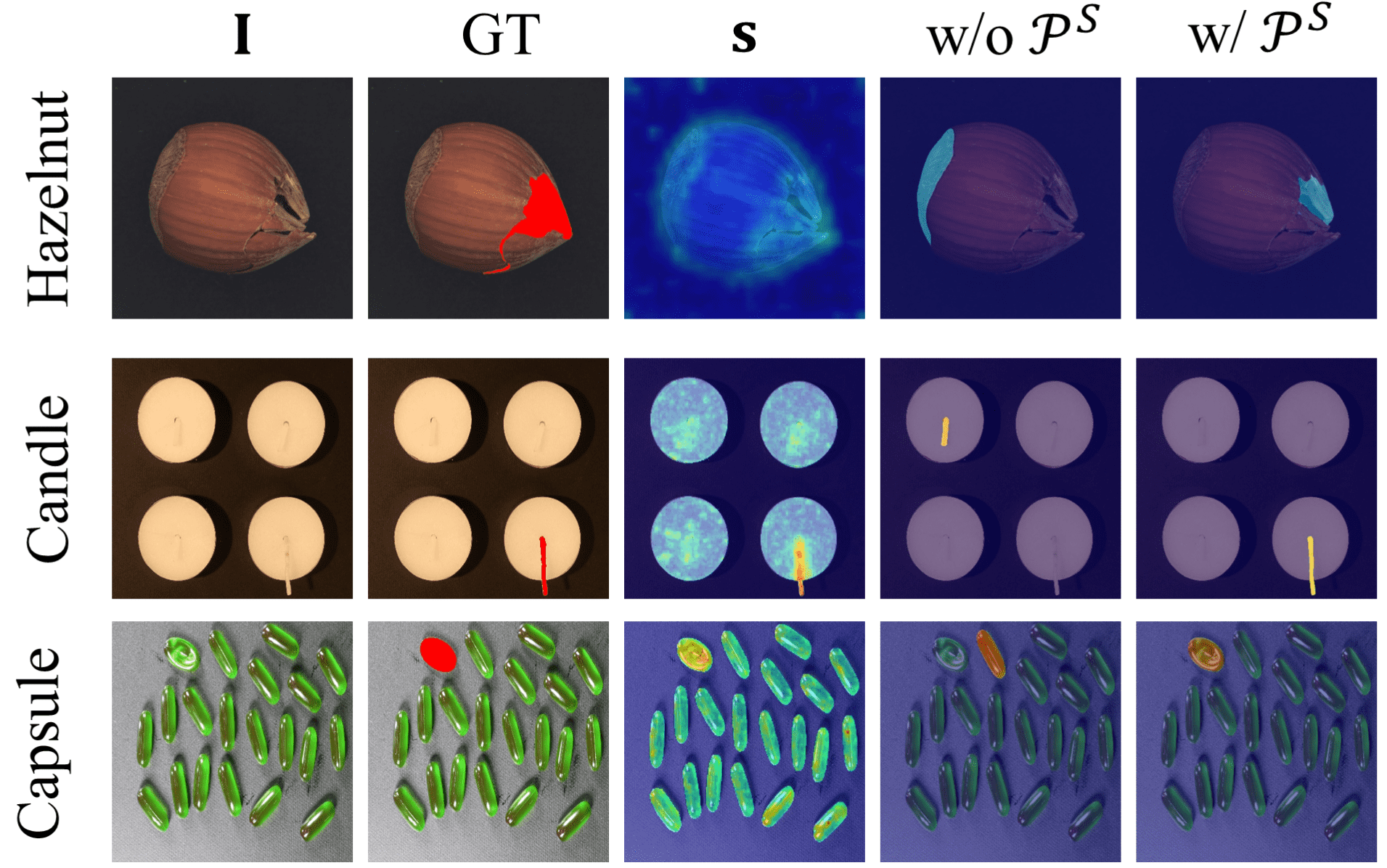}
    \vspace{-6mm}
    \figcaption{Effects of disabling ($w/o$) and abling ($w/$)  prompts ($\mathcal{P}^{S}$) of saliency maps ($\mathbf{s}$) on the final anomaly segmentation.}
  \label{fig:influence_PS} 
  \vspace{2mm}
    \centering
    \includegraphics[width = 0.8\linewidth]{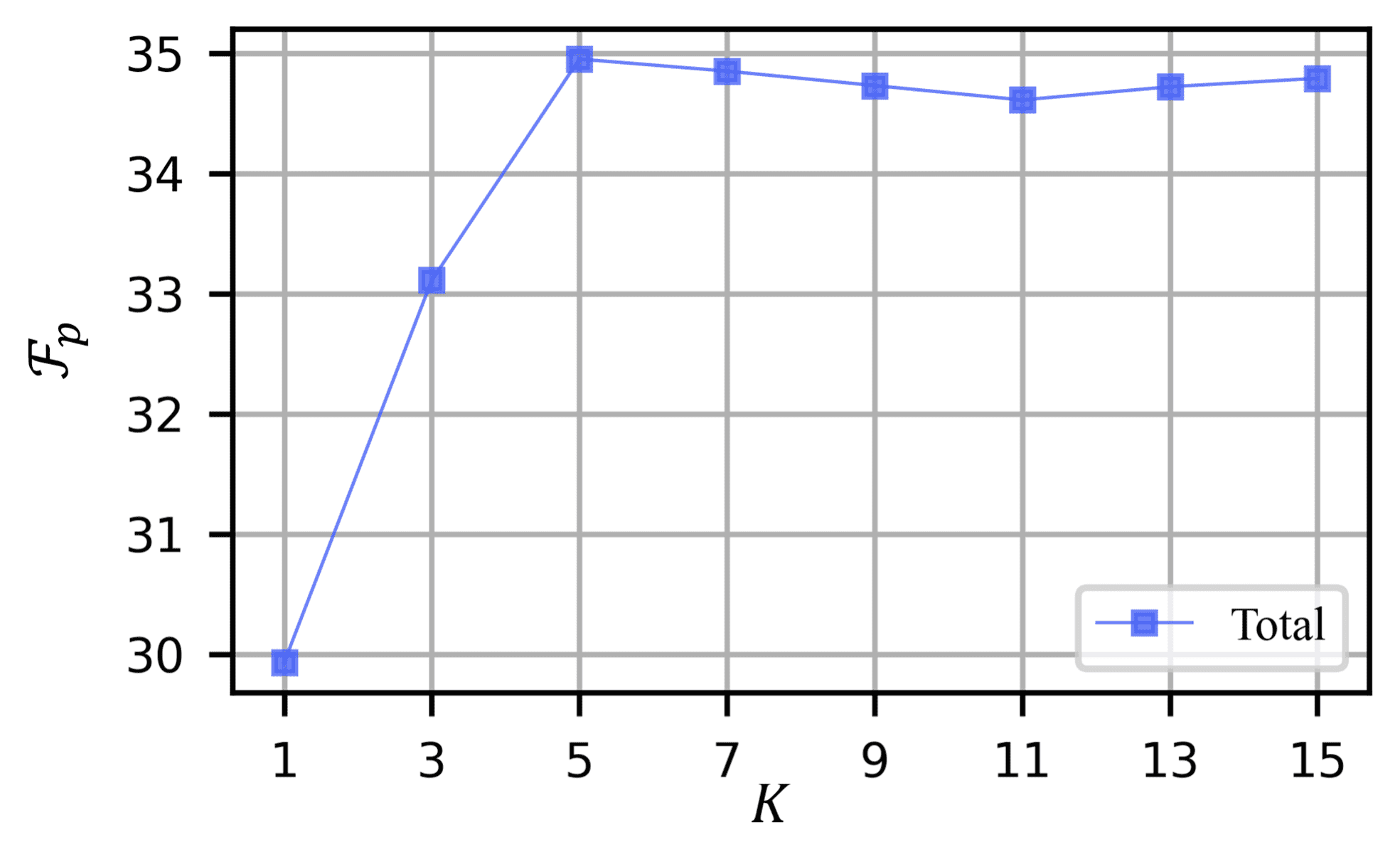}
        \vspace{-1mm}
    \figcaption{Sensitivity analysis of hyperparameter $K$ of confidence prompts ($\mathcal{P}^C$).}
  \label{fig:influence_K}
  \end{minipage}
  \vspace{-2mm}
\end{figtab}

\subsection{Ablation study}
\label{sec:ablation}
In Table \ref{tab:ablation}, we perform component-wise analysis to ablate specific prompt designs in our framework.



\noindent\textbf{Language prompt ($\mathcal{P}^L$).} Table \ref{tab:ablation} verifies the effectiveness of language prompts derived from domain expert knowledge (+3.90\% in $\mathcal{F}_p$ and +4.90\% in $\mathcal{F}_r$). Then, we dig into the efficacy of $\mathcal{T}_{\rm a}$ and $\mathcal{T}_ {\rm s}$, which clearly indicate that both the general description and the specifically designed description for anomalies can achieve reasonable performance. Moreover, their combination can make a synergy, enhancing anomaly segmentation performance. The improvement of $\mathcal{P}^L$ helps unlock language-driven region detection capacity of current foundation models~\cite{liu2023grounding,kirillov2023segment}. 

\noindent\textbf{Property prompt ($\mathcal{P}^P$).} Apart from the improvement in the overall performance, property prompts bring dramatic improvements (from $21.83\%$ to $53.79\%$ in $\mathcal{F}_p$) on \textit{texture} categories, thanks to the filtering mechanism which filters out a significant number of falsely detected anomaly region candidates via high-level characteristics, \textit{e.g.}, location and area of the target image. 

\noindent\textbf{Saliency prompt ($\mathcal{P}^S$).} Table \ref{tab:ablation} provides clear evidence of the efficacy of $\mathcal{P}^S$ on anomaly segmentation. This is because region saliencies can accurately describe the degree of deviation of a region from its surroundings. In Fig. \ref{fig:influence_PS}, we showcase the qualitative impact of $\mathcal{P}^S$ on anomaly segmentation, which illustrates visual saliency maps can help highlight abnormal regions, \textit{i.e.}, which shows higher saliency values compared to other regions. By incorporating $\mathcal{P}^S$ to calibrate the confidence scores, more precise segmentation results can be achieved. For example, the use of $\mathcal{P}^S$ enables the effective localization of the cracked region of hazelnut and the overlong wick on candles. 

\noindent\textbf{Confidence prompt ($\mathcal{P}^C$).} 
 With the incorporation of anomaly confidence prompts, we limit the number of anomaly regions, which effectively reduces false positives, leading to $0.72\%$ $\mathcal{F}_p$ average improvements across all categories, as shown in Table \ref{tab:ablation}.
The influence of the hyperparameter $K$ in $\mathcal{P}^C$ is illustrated in Fig. \ref{fig:influence_K}. The figure shows that performance initially increases as $K$ improves, as more anomaly regions are accurately detected. However, when $K$ exceeds a certain threshold (around $K=5$), the performance drops slightly as more regions are wrongly identified as abnormal. The best results are obtained at around $K=5$, with an average $\mathcal{F}_{p}$ of $34.85\%$  across all categories.

\section{Conclusion}

In this work, we explore how to \textit{segment any anomaly} without any further training by unleashing the full power of modern foundation models. We owe the struggle of adapting foundation model assembly to anomaly segmentation to the prompt design, which is the key to controlling the function of off-the-shelf foundation models. Thus, we propose a novel framework, \textit{i.e.}, Segment Any Anomaly $+$, to leverage hybrid prompts derived from both expert knowledge and target image context to regularize foundation models free of training. Finally, we successfully adapt multiple foundation models to tackle zero-shot anomaly segmentation, achieving new SoTA results on several benchmarks. We hope our work can shed light on the design of label-free model adaptation for anomaly segmentation.

\textbf{Limitations.} Due to the computation restriction, we currently do not test our method on more large-scale foundation models. We have finished the exploration of our methodology with representative foundation models, and we will explore the scaling effect of the models in the future.

{\small


\begin{thebibliography}{10}

\bibitem{cao_collaborative_2023}
Yunkang Cao, Xiaohao Xu, Zhaoge Liu, and Weiming Shen.
\newblock Collaborative discrepancy optimization for reliable image anomaly
  localization.
\newblock {\em IEEE Transactions on Industrial Informatics}, pages 1--10, 2023.

\bibitem{wan_industrial_2022}
Qian Wan, Liang Gao, Xinyu Li, and Long Wen.
\newblock Industrial image anomaly localization based on gaussian clustering of
  pretrained feature.
\newblock {\em IEEE Transactions on Industrial Electronics}, 69(6):6182--6192,
  2021.

\bibitem{roth2022towards}
Karsten Roth, Latha Pemula, Joaquin Zepeda, Bernhard Sch{\"o}lkopf, Thomas
  Brox, and Peter Gehler.
\newblock Towards total recall in industrial anomaly detection.
\newblock In {\em Proceedings of the IEEE/CVF Conference on Computer Vision and
  Pattern Recognition}, pages 14318--14328, 2022.

\bibitem{bergmann2019mvtec}
Paul Bergmann, Michael Fauser, David Sattlegger, and Carsten Steger.
\newblock {MVTec AD} -- {A} comprehensive real-world dataset for unsupervised
  anomaly detection.
\newblock In {\em Proceedings of the IEEE/CVF conference on Computer Vision and
  Pattern Recognition}, pages 9592--9600, 2019.

\bibitem{bergmann2020uninformed}
Paul Bergmann, Michael Fauser, David Sattlegger, and Carsten Steger.
\newblock Uninformed students: Student-teacher anomaly detection with
  discriminative latent embeddings.
\newblock In {\em Proceedings of the IEEE/CVF Conference on Computer Vision and
  Pattern Recognition}, pages 4183--4192, 2020.

\bibitem{baur_autoencoders_2021}
Christoph Baur, Stefan Denner, Benedikt Wiestler, Nassir Navab, and Shadi
  Albarqouni.
\newblock Autoencoders for unsupervised anomaly segmentation in brain mr
  images: a comparative study.
\newblock {\em Medical Image Analysis}, 69:101952, 2021.

\bibitem{zhou2020encoding}
Kang Zhou, Yuting Xiao, Jianlong Yang, Jun Cheng, Wen Liu, Weixin Luo, Zaiwang
  Gu, Jiang Liu, and Shenghua Gao.
\newblock Encoding structure-texture relation with p-net for anomaly detection
  in retinal images.
\newblock In {\em Computer Vision--ECCV 2020: 16th European Conference,
  Glasgow, UK, August 23--28, 2020, Proceedings, Part XX 16}, pages 360--377.
  Springer, 2020.

\bibitem{hou2021divide}
Jinlei Hou, Yingying Zhang, Qiaoyong Zhong, Di~Xie, Shiliang Pu, and Hong Zhou.
\newblock Divide-and-assemble: Learning block-wise memory for unsupervised
  anomaly detection.
\newblock In {\em Proceedings of the IEEE/CVF International Conference on
  Computer Vision}, pages 8791--8800, 2021.

\bibitem{zavrtanik2021draem}
Vitjan Zavrtanik, Matej Kristan, and Danijel Sko{\v{c}}aj.
\newblock {DRAEM} -- {A} discriminatively trained reconstruction embedding for
  surface anomaly detection.
\newblock In {\em Proceedings of the IEEE/CVF International Conference on
  Computer Vision}, pages 8330--8339, 2021.

\bibitem{matsubara2020deep}
Takashi Matsubara, Kazuki Sato, Kenta Hama, Ryosuke Tachibana, and Kuniaki
  Uehara.
\newblock Deep generative model using unregularized score for anomaly detection
  with heterogeneous complexity.
\newblock {\em IEEE Transactions on Cybernetics}, 52(6):5161--5173, 2020.

\bibitem{yan2021learning}
Xudong Yan, Huaidong Zhang, Xuemiao Xu, Xiaowei Hu, and Pheng-Ann Heng.
\newblock Learning semantic context from normal samples for unsupervised
  anomaly detection.
\newblock In {\em Proceedings of the AAAI Conference on Artificial
  Intelligence}, volume~35, pages 3110--3118, 2021.

\bibitem{jiang2022masked}
Jielin Jiang, Jiale Zhu, Muhammad Bilal, Yan Cui, Neeraj Kumar, Ruihan Dou,
  Feng Su, and Xiaolong Xu.
\newblock Masked swin transformer unet for industrial anomaly detection.
\newblock {\em IEEE Transactions on Industrial Informatics}, 19(2):2200--2209,
  2022.

\bibitem{yi2020patch}
Jihun Yi and Sungroh Yoon.
\newblock Patch {SVDD}: Patch-level {SVDD} for anomaly detection and
  segmentation.
\newblock In {\em Proceedings of the Asian Conference on Computer Vision},
  2020.

\bibitem{massoli2021mocca}
Fabio~Valerio Massoli, Fabrizio Falchi, Alperen Kantarci, {\c{S}}eymanur Akti,
  Hazim~Kemal Ekenel, and Giuseppe Amato.
\newblock Mocca: Multilayer one-class classification for anomaly detection.
\newblock {\em IEEE Transactions on Neural Networks and Learning Systems},
  33(6):2313--2323, 2021.

\bibitem{sohn2020learning}
Kihyuk Sohn, Chun-Liang Li, Jinsung Yoon, Minho Jin, and Tomas Pfister.
\newblock Learning and evaluating representations for deep one-class
  classification.
\newblock In {\em International Conference on Learning Representations}, 2020.

\bibitem{cao2023complementary}
Yunkang Cao, Xiaohao Xu, and Weiming Shen.
\newblock Complementary pseudo multimodal feature for point cloud anomaly
  detection.
\newblock {\em arXiv preprint arXiv:2303.13194}, 2023.

\bibitem{wang_multimodal_nodate}
Yue Wang, Jinlong Peng, Jiangning Zhang, Ran Yi, Yabiao Wang, and Chengjie
  Wang.
\newblock Multimodal industrial anomaly detection via hybrid fusion.
\newblock In {\em 2023 {IEEE}/{CVF} Conference on Computer Vision and Pattern
  Recognition ({CVPR})}, 2023.

\bibitem{jiang_softpatch_2022}
Xi~Jiang, Jianlin Liu, Jinbao Wang, Qiang Nie, Kai Wu, Yong Liu, Chengjie Wang,
  and Feng Zheng.
\newblock {SoftPatch}: Unsupervised anomaly detection with noisy data.
\newblock In {\em Advances in neural information processing systems}, 2022.

\bibitem{kirillov2023segment}
Alexander Kirillov, Eric Mintun, Nikhila Ravi, Hanzi Mao, Chloe Rolland, Laura
  Gustafson, Tete Xiao, Spencer Whitehead, Alexander~C Berg, Wan-Yen Lo, et~al.
\newblock Segment anything.
\newblock {\em arXiv preprint arXiv:2304.02643}, 2023.

\bibitem{radford2021learning}
Alec Radford, Jong~Wook Kim, Chris Hallacy, Aditya Ramesh, Gabriel Goh,
  Sandhini Agarwal, Girish Sastry, Amanda Askell, Pamela Mishkin, Jack Clark,
  et~al.
\newblock Learning transferable visual models from natural language
  supervision.
\newblock In {\em International Conference on Machine Learning}, pages
  8748--8763. PMLR, 2021.

\bibitem{li2022align}
Dongxu Li, Junnan Li, Hongdong Li, Juan~Carlos Niebles, and Steven~CH Hoi.
\newblock Align and prompt: Video-and-language pre-training with entity
  prompts.
\newblock In {\em Proceedings of the IEEE/CVF Conference on Computer Vision and
  Pattern Recognition}, pages 4953--4963, 2022.

\bibitem{bommasani2021opportunities}
Rishi Bommasani, Drew~A Hudson, Ehsan Adeli, Russ Altman, Simran Arora, Sydney
  von Arx, Michael~S Bernstein, Jeannette Bohg, Antoine Bosselut, Emma
  Brunskill, et~al.
\newblock On the opportunities and risks of foundation models.
\newblock {\em arXiv preprint arXiv:2108.07258}, 2021.

\bibitem{liu2023grounding}
Shilong Liu, Zhaoyang Zeng, Tianhe Ren, Feng Li, Hao Zhang, Jie Yang, Chunyuan
  Li, Jianwei Yang, Hang Su, Jun Zhu, et~al.
\newblock Grounding dino: Marrying dino with grounded pre-training for open-set
  object detection.
\newblock {\em arXiv preprint arXiv:2303.05499}, 2023.

\bibitem{clipseg2022}
Timo L{\"u}ddecke and Alexander Ecker.
\newblock Image segmentation using text and image prompts.
\newblock In {\em Proceedings of the IEEE/CVF Conference on Computer Vision and
  Pattern Recognition}, pages 7086--7096, 2022.

\bibitem{jeong2023winclip}
Jongheon Jeong, Yang Zou, Taewan Kim, Dongqing Zhang, Avinash Ravichandran, and
  Onkar Dabeer.
\newblock Winclip: Zero-/few-shot anomaly classification and segmentation.
\newblock {\em arXiv preprint arXiv:2303.14814}, 2023.

\bibitem{object_calibration}
Xiaohao Xu, Jinglu Wang, Xiang Ming, and Yan Lu.
\newblock Towards robust video object segmentation with adaptive object
  calibration.
\newblock In {\em Proceedings of the 30th ACM International Conference on
  Multimedia}, pages 1--10, 2022.

\bibitem{xu2022reliable}
Xiaohao Xu, Jinglu Wang, Xiao Li, and Yan Lu.
\newblock Reliable propagation-correction modulation for video object
  segmentation.
\newblock In {\em Proceedings of the AAAI Conference on Artificial
  Intelligence}, pages 2946--2954, 2022.

\bibitem{paiss_count_2023}
Roni Paiss, Ariel Ephrat, Omer Tov, Shiran Zada, Inbar Mosseri, Michal Irani,
  and Tali Dekel.
\newblock Teaching clip to count to ten.
\newblock {\em arXiv preprint arXiv:2302.12066}, 2023.

\bibitem{li2022r}
Xiang Li, Jinglu Wang, Xiaohao Xu, Xiao Li, Yan Lu, and Bhiksha Raj.
\newblock R\^{} 2vos: Robust referring video object segmentation via relational
  multimodal cycle consistency.
\newblock {\em arXiv preprint arXiv:2207.01203}, 2022.

\bibitem{salehi2021multiresolution}
Mohammadreza Salehi, Niousha Sadjadi, Soroosh Baselizadeh, Mohammad~H Rohban,
  and Hamid~R Rabiee.
\newblock Multiresolution knowledge distillation for anomaly detection.
\newblock In {\em Proceedings of the IEEE/CVF Conference on Computer Vision and
  Pattern Recognition}, pages 14902--14912, 2021.

\bibitem{wang2021student}
Guodong Wang, Shumin Han, Errui Ding, and Di~Huang.
\newblock Student-teacher feature pyramid matching for anomaly detection.
\newblock {\em arXiv preprint arXiv:2103.04257}, 2021.

\bibitem{deng2022anomaly}
Hanqiu Deng and Xingyu Li.
\newblock Anomaly detection via reverse distillation from one-class embedding.
\newblock In {\em Proceedings of the IEEE/CVF Conference on Computer Vision and
  Pattern Recognition}, pages 9737--9746, 2022.

\bibitem{cao2022informative}
Yunkang Cao, Qian Wan, Weiming Shen, and Liang Gao.
\newblock Informative knowledge distillation for image anomaly segmentation.
\newblock {\em Knowledge-Based Systems}, 248:108846, 2022.

\bibitem{cao2022semikd}
Yunkang Cao, Yanan Song, Xiaohao Xu, Shuya Li, Yuhao Yu, Yiheng Zhang, and
  Weiming Shen.
\newblock Semi-supervised knowledge distillation for tiny defect detection.
\newblock In {\em 2022 IEEE 25th International Conference on Computer Supported
  Cooperative Work in Design (CSCWD)}, pages 1010--1015, 2022.

\bibitem{wan_unsupervised_2022}
Qian Wan, Liang Gao, Xinyu Li, and Long Wen.
\newblock Unsupervised image anomaly detection and segmentation based on
  pre-trained feature mapping.
\newblock {\em IEEE Transactions on Industrial Informatics}, 2022.

\bibitem{wan_position_2022}
Qian Wan, Yunkang Cao, Liang Gao, Weiming Shen, and Xinyu Li.
\newblock Position encoding enhanced feature mapping for image anomaly
  detection.
\newblock In {\em 2022 {IEEE} 18th International Conference on Automation
  Science and Engineering ({CASE})}, pages 876--881. {IEEE}, 2022.

\bibitem{nagy2022zero}
Amr~M Nagy and L{\'a}szl{\'o} Cz{\'u}ni.
\newblock Zero-shot learning and classification of steel surface defects.
\newblock In {\em Fourteenth International Conference on Machine Vision (ICMV
  2021)}, volume 12084, pages 386--394. SPIE, 2022.

\bibitem{liu2021zero}
Jiahui Liu, Xiaojuan Qi, Songzhi Su, Tony Prescott, and Li~Sun.
\newblock Zero-shot anomalous object detection using unsupervised metric
  learning.
\newblock In {\em 2021 IEEE/RSJ International Conference on Intelligent Robots
  and Systems (IROS 2021) Proceedings}. Sheffield, 2021.

\bibitem{rivera2020anomaly}
Ad{\'\i}n~Ram{\'\i}rez Rivera, Adil Khan, Imad Eddine~Ibrahim Bekkouch, and
  Taimoor~Shakeel Sheikh.
\newblock Anomaly detection based on zero-shot outlier synthesis and
  hierarchical feature distillation.
\newblock {\em IEEE Transactions on Neural Networks and Learning Systems},
  33(1):281--291, 2020.

\bibitem{aota2023zero}
Toshimichi Aota, Lloyd Teh~Tzer Tong, and Takayuki Okatani.
\newblock Zero-shot versus many-shot: Unsupervised texture anomaly detection.
\newblock In {\em Proceedings of the IEEE/CVF Winter Conference on Applications
  of Computer Vision}, pages 5564--5572, 2023.

\bibitem{Laion400}
Christoph Schuhmann, Robert Kaczmarczyk, Aran Komatsuzaki, Aarush Katta,
  Richard Vencu, Romain Beaumont, Jenia Jitsev, Theo Coombes, and Clayton
  Mullis.
\newblock Laion-400m: Open dataset of clip-filtered 400 million image-text
  pairs.
\newblock In {\em NeurIPS Workshop Datacentric AI}. J{\"u}lich Supercomputing
  Center, 2021.

\bibitem{li2021align}
Junnan Li, Ramprasaath Selvaraju, Akhilesh Gotmare, Shafiq Joty, Caiming Xiong,
  and Steven Chu~Hong Hoi.
\newblock Align before fuse: Vision and language representation learning with
  momentum distillation.
\newblock In {\em Advances in neural information processing systems},
  volume~34, pages 9694--9705, 2021.

\bibitem{lu2022unified}
Jiasen Lu, Christopher Clark, Rowan Zellers, Roozbeh Mottaghi, and Aniruddha
  Kembhavi.
\newblock {Unified-IO}: A unified model for vision, language, and multi-modal
  tasks.
\newblock {\em arXiv preprint arXiv:2206.08916}, 2022.

\bibitem{wang2022unifying}
Peng Wang, An~Yang, Rui Men, Junyang Lin, Shuai Bai, Zhikang Li, Jianxin Ma,
  Chang Zhou, Jingren Zhou, and Hongxia Yang.
\newblock Unifying architectures, tasks, and modalities through a simple
  sequence-to-sequence learning framework.
\newblock {\em arXiv preprint arXiv:2202.03052}, 2022.

\bibitem{rao2022denseclip}
Yongming Rao, Wenliang Zhao, Guangyi Chen, Yansong Tang, Zheng Zhu, Guan Huang,
  Jie Zhou, and Jiwen Lu.
\newblock Denseclip: Language-guided dense prediction with context-aware
  prompting.
\newblock In {\em Proceedings of the IEEE Conference on Computer Vision and
  Pattern Recognition (CVPR)}, 2022.

\bibitem{zhong2022regionclip}
Yiwu Zhong, Jianwei Yang, Pengchuan Zhang, Chunyuan Li, Noel Codella,
  Liunian~Harold Li, Luowei Zhou, Xiyang Dai, Lu~Yuan, Yin Li, et~al.
\newblock Regionclip: Region-based language-image pretraining.
\newblock In {\em Proceedings of the IEEE/CVF Conference on Computer Vision and
  Pattern Recognition}, pages 16793--16803, 2022.

\bibitem{zhou2021denseclip}
Chong Zhou, Chen~Change Loy, and Bo~Dai.
\newblock Extract free dense labels from clip.
\newblock In {\em Computer Vision--ECCV 2022: 17th European Conference, Tel
  Aviv, Israel, October 23--27, 2022, Proceedings, Part XXVIII}, pages
  696--712. Springer, 2022.

\bibitem{zhou_conditional}
Kaiyang Zhou, Jingkang Yang, Chen~Change Loy, and Ziwei Liu.
\newblock Conditional prompt learning for vision-language models.
\newblock In {\em 2022 {IEEE}/{CVF} Conference on Computer Vision and Pattern
  Recognition ({CVPR})}, 2022.

\bibitem{ju_prompting_2022}
Chen Ju, Tengda Han, Kunhao Zheng, Ya~Zhang, and Weidi Xie.
\newblock Prompting visual-language models for efficient video understanding.
\newblock In {\em Computer Vision--ECCV 2022: 17th European Conference, Tel
  Aviv, Israel, October 23--27, 2022, Proceedings, Part XXXV}, pages 105--124.
  Springer, 2022.

\bibitem{jia_visual_2022}
Menglin Jia, Luming Tang, Bor-Chun Chen, Claire Cardie, Serge Belongie, Bharath
  Hariharan, and Ser-Nam Lim.
\newblock Visual prompt tuning.
\newblock In {\em Computer Vision--ECCV 2022: 17th European Conference, Tel
  Aviv, Israel, October 23--27, 2022, Proceedings, Part XXXIII}, pages
  709--727. Springer, 2022.

\bibitem{bahng_exploring_2022}
Hyojin Bahng, Ali Jahanian, Swami Sankaranarayanan, and Phillip Isola.
\newblock Exploring visual prompts for adapting large-scale models.
\newblock {\em arXiv preprint arXiv:2203.17274}, 1(3):4, 2022.

\bibitem{zang_unified_2022}
Yuhang Zang, Wei Li, Kaiyang Zhou, Chen Huang, and Chen~Change Loy.
\newblock Unified vision and language prompt learning.
\newblock {\em arXiv preprint arXiv:2210.07225}, 2022.

\bibitem{shen_multitask_2022}
Sheng Shen, Shijia Yang, Tianjun Zhang, Bohan Zhai, Joseph~E Gonzalez, Kurt
  Keutzer, and Trevor Darrell.
\newblock Multitask vision-language prompt tuning.
\newblock {\em arXiv preprint arXiv:2211.11720}, 2022.

\bibitem{zhou_learning_2022}
Kaiyang Zhou, Jingkang Yang, Chen~Change Loy, and Ziwei Liu.
\newblock Learning to prompt for vision-language models.
\newblock {\em Int J Comput Vis}, 130(9):2337--2348, 2022.

\bibitem{shtedritski_what_2023}
Aleksandar Shtedritski, Christian Rupprecht, and Andrea Vedaldi.
\newblock What does clip know about a red circle? visual prompt engineering for
  vlms.
\newblock {\em arXiv preprint arXiv:2304.06712}, 2023.

\bibitem{dosovitskiy2021an}
Alexey Dosovitskiy, Lucas Beyer, Alexander Kolesnikov, Dirk Weissenborn,
  Xiaohua Zhai, Thomas Unterthiner, Mostafa Dehghani, Matthias Minderer, Georg
  Heigold, Sylvain Gelly, Jakob Uszkoreit, and Neil Houlsby.
\newblock An image is worth 16x16 words: Transformers for image recognition at
  scale.
\newblock In {\em International Conference on Learning Representations}, 2021.

\bibitem{li2021grounded}
Liunian~Harold Li, Pengchuan Zhang, Haotian Zhang, Jianwei Yang, Chunyuan Li,
  Yiwu Zhong, Lijuan Wang, Lu~Yuan, Lei Zhang, Jenq-Neng Hwang, et~al.
\newblock Grounded language-image pre-training.
\newblock In {\em Proceedings of the IEEE/CVF Conference on Computer Vision and
  Pattern Recognition}, pages 10965--10975, 2022.

\bibitem{zou2022spot}
Yang Zou, Jongheon Jeong, Latha Pemula, Dongqing Zhang, and Onkar Dabeer.
\newblock {SPot-the-Difference} self-supervised pre-training for anomaly
  detection and segmentation.
\newblock In {\em Proceedings of the European Conference on Computer Vision},
  2022.

\bibitem{hinton2012imagenet}
Geoffrey~E Hinton, Alex Krizhevsky, and Ilya Sutskever.
\newblock Image{N}et classification with deep convolutional neural networks.
\newblock {\em Advances in Neural Information Processing Systems},
  25(1106-1114):1, 2012.

\bibitem{bozic_mixed_2021}
Jakob Bo{\v{z}}i{\v{c}}, Domen Tabernik, and Danijel Sko{\v{c}}aj.
\newblock Mixed supervision for surface-defect detection: From weakly to fully
  supervised learning.
\newblock {\em Computers in Industry}, 129:103459, 2021.

\bibitem{mtd2018}
Yibin Huang, Congying Qiu, Yue Guo, Xiaonan Wang, and Kui Yuan.
\newblock Surface defect saliency of magnetic tile.
\newblock In {\em 2018 IEEE 14th International Conference on Automation Science
  and Engineering (CASE)}, pages 612--617, 2018.

\bibitem{zagoruyko2016wideresnet}
Sergey Zagoruyko and Nikos Komodakis.
\newblock Wide residual networks.
\newblock In Edwin R.~Hancock Richard C.~Wilson and William A.~P. Smith,
  editors, {\em Proceedings of the British Machine Vision Conference (BMVC)},
  pages 87.1--87.12. BMVA Press, September 2016.

\end{thebibliography}

}


\end{document}